\documentclass[letterpaper, 10 pt, conference]{ieeeconf}  
\pdfoutput=1

\IEEEoverridecommandlockouts                              

\usepackage{times}
\usepackage{epsfig}
\usepackage{graphicx}
\usepackage{amsmath}
\usepackage{amssymb}
\usepackage{booktabs}
\usepackage{multirow}
\usepackage{widetable}

\title{\LARGE \bf
Associating Grasp Configurations with Hierarchical Features in Convolutional Neural Networks
}

\author{Li Yang Ku, Erik Learned-Miller, and Rod Grupen
\thanks{Li Yang Ku, Erik Learned-Miller, and Rod Grupen are with the College of Information and Computer Sciences, University of Massachusetts Amherst,
        Amherst, MA 01003, USA
        {\tt\small \{lku,elm,grupen\}@cs.umass.edu }}%
}

\begin{document}

\maketitle
\thispagestyle{empty}
\pagestyle{empty}

\begin{abstract}
In this work, we provide a solution for posturing the anthropomorphic Robonaut-2 hand and arm for grasping based on visual information. A mapping from visual features extracted from a convolutional neural network (CNN) to grasp points is learned. We demonstrate that a CNN pre-trained for image classification can be applied to a grasping task based on a small set of grasping examples. Our approach takes advantage of the hierarchical nature of the CNN by identifying features that capture the hierarchical support relations between filters in different CNN layers and locating their 3D positions by tracing activations backwards in the CNN. When this backward trace terminates in the RGB-D image, important manipulable structures are thereby localized. These features that reside in different layers of the CNN are then associated with controllers that engage different kinematic subchains in the hand/arm system for grasping. A grasping dataset is collected using demonstrated hand/object relationships for Robonaut-2 to evaluate the proposed approach in terms of the precision of the resulting preshape postures. We demonstrate that this approach outperforms baseline approaches in cluttered scenarios on the grasping dataset and a point cloud based approach on a grasping task using \mbox{Robonaut-2}.

\end{abstract}

\section{INTRODUCTION}

In this work, we provide a solution for posturing Robonaut-2's anthropomorphic hand and arm for grasping based on visual information. Experiments conducted by Goodale and Milner \cite{goodale2013sight} have shown that humans are capable of pre-shaping their hands to grasp an object based exclusively on visual information from an object. 
There are, in general, many possible kinds of grasps for each object; here we will focus on generating a specific grasp that is similar to a demonstrated grasp. 

\begin{figure}[!ht]
\centering
\includegraphics[width=1.\linewidth]{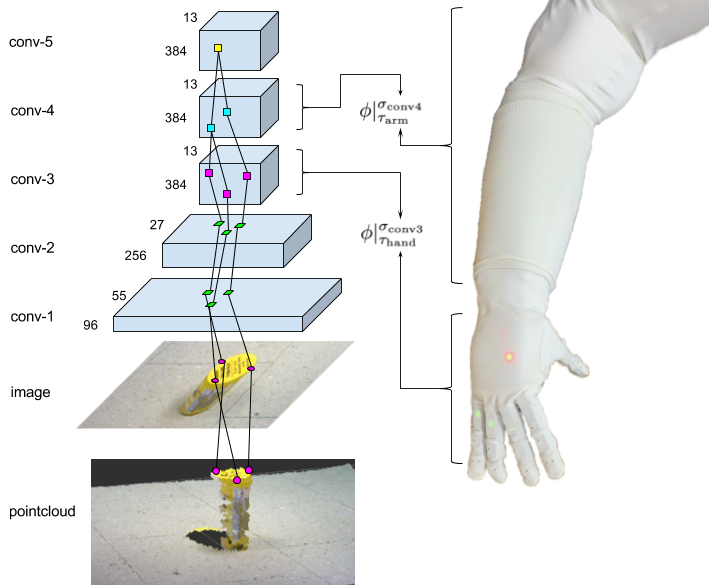}
\caption{Overall architecture of the proposed system. The input for this example is the RGB image and point cloud of a yellow jar. Tuples of the yellow dots (conv-5), the cyan dots (conv-4), and the magenta dots (conv-3) represent hierarchical CNN features. 
These features can be traced back to the image input and mapped to the point cloud to support manipulation. $\mathrm{\phi |^{\sigma_{conv3}}_{\tau_{hand}}}$ represents the function that controls hand motor resources $\mathrm{\tau_{hand}}$ based on conv-3 layer features $\mathrm{\sigma_{conv3}}$ and $\mathrm{\phi |^{\sigma_{conv4}}_{\tau_{arm}}}$ represents the function that controls arm motor resources $\mathrm{\tau_{arm}}$ based on conv-4 layer features $\mathrm{\sigma_{conv4}}$. $\phi$ represents the potential function computed from input $\sigma$ whose derivative with respect to motor resources $\tau$ constitute a controller.
}
\label{fig:archi}
\end{figure}

Convolutional neural networks (CNNs) have attracted a great deal of attention in the computer vision community and have outperformed other algorithms on many benchmarks. However, applying CNNs to robotics applications is non-trivial for two reasons. First, collecting the quantity of robot data typically required to train a CNN is extremely difficult. 
In this work, we learn a mapping from visual features extracted from a CNN trained on ImageNet \cite{ILSVRC15} to grasp configurations of the entire anthropomorphic hand/arm and demonstrate that a small set of grasping examples is sufficient for generalizing grasps across different objects of similar shapes. Second, the final output of a CNN contains little location information from the observed object, which is essential for grasping. 
We take advantage of the hierarchical nature of the CNN to identify a set of features we call \textit{hierarchical CNN features} that capture the hierarchical support relations between filters in different CNN layers. The 3D positions of such features can be identified by tracing activations of high-level filters backwards in the CNN to discover the locations of important structures that direct hand/arm control. Figure \ref{fig:archi} shows the overall architecture and how such hierarchical CNN features are mapped to a point cloud to support manipulation.

We collected a grasping dataset using demonstrated hand/object relationships for Robonaut-2 \cite{diftler2011robonaut} to evaluate the proposed approach in terms of the precision of the resulting preshape postures. 
We show that this proposed approach outperforms alternative approaches in cluttered scenarios on this dataset.

\begin{figure}[thb]
\centering
\includegraphics[width=1\linewidth]{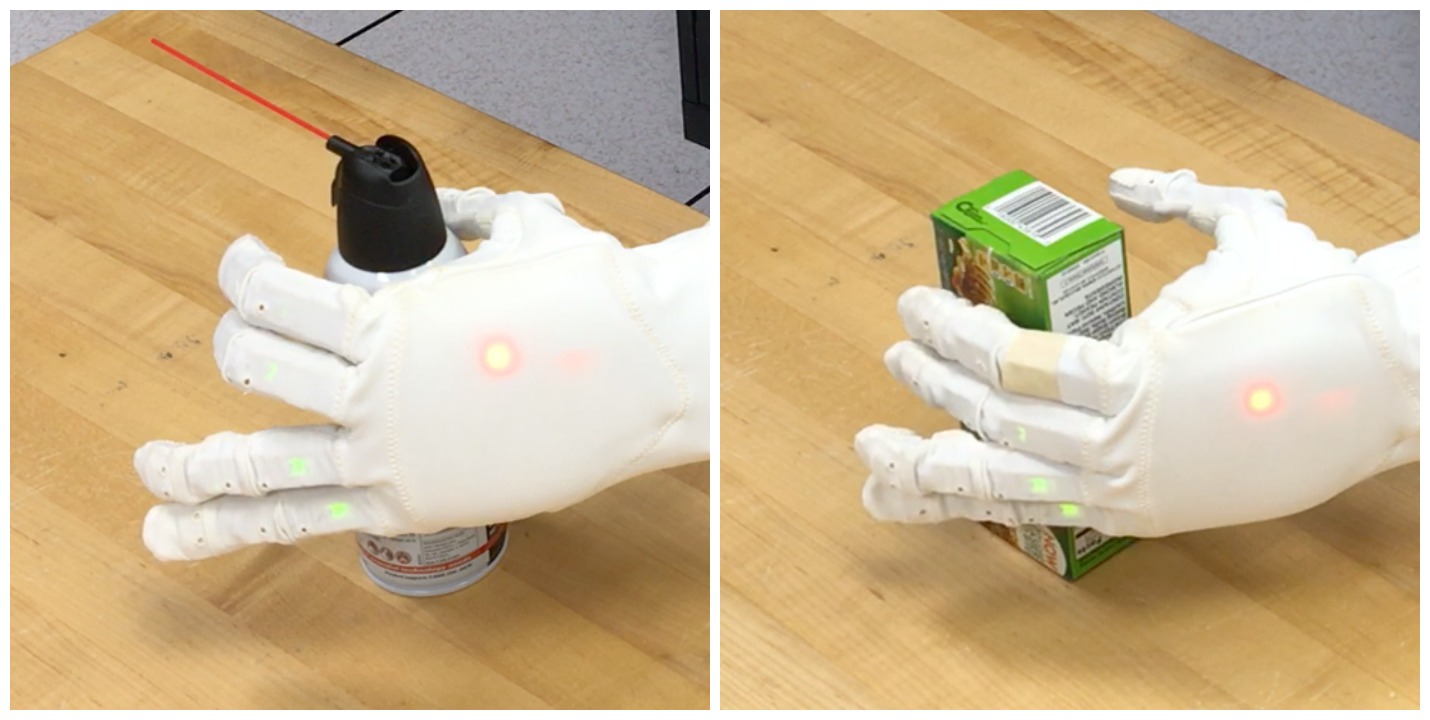}
\caption{Robonaut-2 pre-shaping its hand before grasping an object. Notice that in the air duster example the fingers are pre-shaped to form an enveloping grasp and in the box example the fingers are pre-shaped to grasp the faces of the box. }
\label{fig:preshape}
\end{figure}

This approach is then tested in a grasping experiment on Robonaut-2 which hierarchical CNN features in the 4th and 5th layers are associated with the arm and hand controllers respectively.
To our knowledge, this is the first work that associates features in different CNN layers with controllers that engage different kinematic subchains in the hand/arm systems. We demonstrate that the proposed approach outperforms a baseline approach using the object point cloud. Figure \ref{fig:preshape} shows Robonaut-2 pre-shaping its hand before grasping novel cylindrical and cuboid objects.

\section{RELATED WORK}
The idea that our brain encodes visual stimuli in two separate regions was first proposed by Schneider \cite{schneider1969two}.
Ungerleider and Mishkin further discovered the ventral and dorsal streams and proposed the hypothesis often known as the distinction of ``what" and ``where" between the two visual pathways \cite{mishkin1983object}.  
However, in 1992 Goodale and Milner proposed an alternative perspective on the functionality of these two visual pathways based on many observations made with patient DF \cite{goodale1992separate}. Patient DF developed a profound visual form agnosia due to damage to her ventral stream. Despite DF's inability to recognize the shape, size and orientation of visual objects, she is capable of grasping the object with accurate hand and finger movements. Based on a series of experiments \cite{goodale2013sight}, Goodale and Milner proposed the \textit{perception-action model}, which suggests that the dorsal pathway provides action-relevant information about the structural characteristic of objects in addition to their position. 
Our work is inspired by these observations and associates grasp configurations with visual features instead of object identities.

Recently, studies on this perception-action model in the field of cognitive science have shifted from identifying functional dissociation between ventral and dorsal streams and, instead, focus on how these two streams interact \cite{mcintosh2009two}. Attentional blink experiments \cite{adamo2009picture} suggest that object recognition processes in the ventral stream activate action-related processing in the dorsal stream, which leads to increased attention to objects that present an affordance relationship with the observed object. This interaction between the ventral and dorsal streams disappears when the stimuli are words rather than images. This result is consistent with our proposition that features learned for recognition tasks provide useful visual cues for grasp actions.

A large body of literature has focused on solving for grasp configurations. Sahbani \textit{et al}. divide grasp synthesis approaches into analytic and empirical   \cite{sahbani2012overview} while Bohg \textit{et al}. further categorize grasp methodologies based on prior object knowledge, object feature, task, etc.~\cite{bohg2014data}. Our work belongs to a set of empirical approaches that try to grasp familiar objects using both 2D and 3D features based on demonstrations. 
We compare our work to research that generates robot grasps from visual information.
In work by Saxena \textit{et al}., a single grasp point is identified using a probabilistic model conditioned on a set of visual features such as edges, textures, and colors \cite{saxena2008robotic}. Similar work uses contact, center of mass, and force closure properties based on point cloud and image information to calculate the probability of a hand configuration successfully grasping a novel object \cite{saxena2008learning}.  Platt \textit{et al}. uses online learning to associate different types of grasps with the object's height and width \cite{platt2005re}. A shape template approach for grasping novel objects was also proposed \cite{herzog2014learning}. The authors introduced a shape descriptor called a height map to capture local object geometry for matching a point cloud generated by a novel object to a known grasp template. Other work uses a geometric approach for grasping novel objects based on point clouds \cite{pas2015using}. An antipodal grasp is determined by finding cutting planes that satisfy geometric constraints. A similar approach based on local object geometry was also introduced \cite{zhang2011grasp}.  

Our approach associates CNN features trained on the ImageNet dataset with a demonstrated grasp. Hierarchical features consider both the local pattern and higher level structures that it belongs to. 
This visually-guided grasp can be successful even when there is insufficient 3D point cloud data. For example, a side grasp on a cylinder can be inferred even when only the top face is observed.
In \cite{lenz2015deep}, a deep network trained on 1035 examples is used to determine a successful grasp based on RGB-D data. Grasp positions are exhaustively generated and evaluated. Our approach localizes features in a pre-trained CNN and generates grasp points based on a small set of grasping examples. 

Several authors have applied CNNs to robotics. In \cite{levine2015end}, visuomotor policies are learned using an end-to-end neural network that takes images and outputs joint torques. A three layer CNN is used without any max pooling layer to maintain spatial information. In our work, we also use filters in the third convolution layers; but unlike the previous work, we consider their relationship with higher layer filters. In \cite{finn2015deep}, an autoencoder is used to learn spatial information of features in a neural network. Our approach finds the multi-level receptive field of a hierarchical feature in a particular image by repeatedly back tracing along a single path. In \cite{pinto2015supersizing}, a CNN is used to learn what features are graspable through 50 thousand trials collected using a Baxter robot. The final layer is used to select 1 out of 18 grasp orientations. In contrast, our approach considers multi-objective configurations capable of control for more sophisticated and higher degree-of-freedom hand/arm systems like Robonaut-2. 

A great deal of research has been done on understanding the relationship between CNN filter activation and the input image. In \cite{zeiler2014visualizing}, 
deconvolution is used to find what pixels activate each filter. 
In other work, the gradient of the network response with respect to the image is calculated to obtain a saliency map used for object localization \cite{simonyan2013deep}. In \cite{springenberg2014striving}, an approach that adds a guidance signal to backpropagation for better visualization of higher level filters is also introduced. In our work, backpropagation is performed on a single filter per layer to consider the hierarchical relationship between filters. Recent work by Zhang \textit{et al}. introduces the excitation backprop that uses a probabilistic winner-take-all process to generate attention maps for different categories \cite{zhang2016top}. Our work localizes features based on similar concepts.


Some authors have explored using intermediate filter activation in addition to the the response of the output layer of a CNN. Hypercolumns, which are defined as the activation of all CNN units above a pixel, are used on tasks such as simultaneous detection and segmentation, keypoint localization, and part labelling \cite{hariharan2015hypercolumns}. Our approach groups filters in different layers based on their hierarchical activation instead of just the spatial relationship. In \cite{schwarz2015rgb}, the last two layers of two CNNs, one that takes an image as input and one that takes depth as input, are used to identify object category, instance, and pose. In \cite{wilkinson2015efficient}, the last layer is used to identify the object instance while the fifth convolution layer is used to determine the \textit{aspect} of an object. 
In our work we consider a feature as the activation of a lower layer filter that causes a specific higher layer filter to activate and plan grasp poses based on these features.

Our work is also inspired by \textit{deformable part models} \cite{felzenszwalb2010object} where a class model is composed of smaller models of parts; e.g., wheels are parts of a bicycle. We view CNNs similarly; 
if a higher layer filter response represents a high level structure, the lower layer filter responses that contribute to this higher layer response can be seen as representing local parts of this structure and may provide useful information for manipulating an object.

\begin{figure}[thb]
\centering
\includegraphics[width=0.8\linewidth]{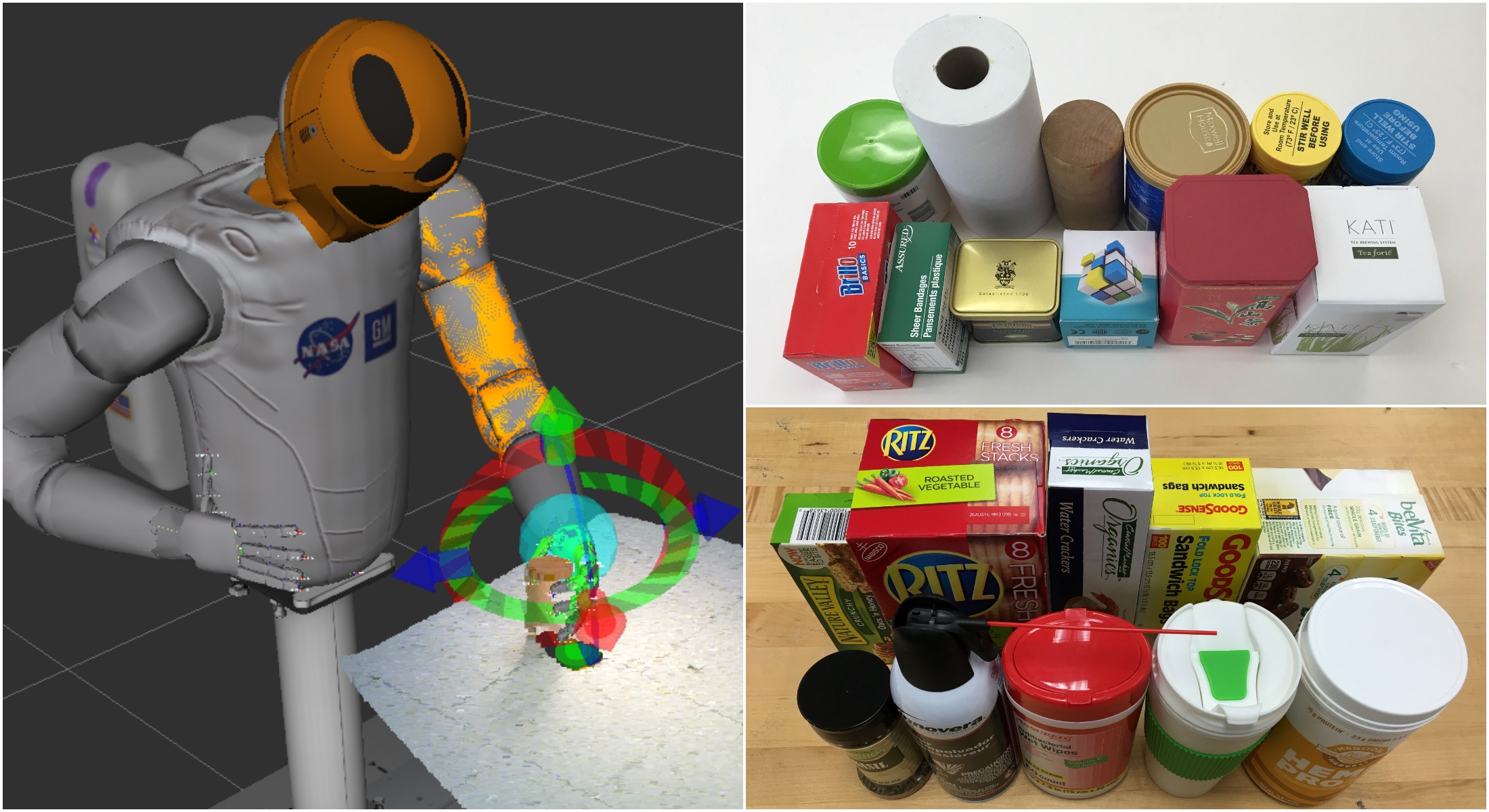}
\caption{Left: the data collection interface where the robot arm and hand is adjusted to the grasp pose. Right top: The set of objects used in the R2 grasping dataset. Right bottom: The set of novel objects used in the grasping experiment.}
\vspace{-0.7em}
\label{fig:dataset}
\end{figure}

\section{DATASET}

In this work, we created the R2 grasping dataset that contains grasp records of each demonstrated grasp. A grasp record contains the point cloud and RGB image of the target object observed from the robot's viewpoint, and the Cartesian pose of each joint in the Robonaut-2 hand in the camera frame. The data is collected using an Asus Xtion camera and the Robonaut-2 simulator \cite{r2sim}. The object is placed on a flat surface  where the camera is about 70 cm above and looking down at a 55 degree angle. 
We manually adjust the left robot arm and each finger of the left hand so that the robot hand can perform a firm grasp on the object. For cuboid objects, we adjust the thumb tip and index finger tip to the front and back faces of the cuboid and about 3cm away from the left edge of the face. For cylindrical objects, we adjust the thumb tip and index finger to perform an enveloping grasp on the object. 


We collected a total of 120 grasping examples of twelve different objects. Six of the objects are cylindrical and six of them are cuboids as shown in Figure \ref{fig:dataset}. The same object is presented at different orientations and under different lighting conditions. 
In addition to grasping examples with a single object, we also created 24 grasping examples in cluttered scenarios. Twelve of them include a single cylindrical object and twelve of them include a single cuboid object. The joint poses of the hand while grasping these objects are also recorded. Figure \ref{fig:clutter} shows a few examples of the cluttered test set.

\section{APPROACH}

To learn how to grasp a previously unseen object, we use a CNN model similar to Alexnet \cite{krizhevsky2012imagenet} introduced by Yosinski \cite{yosinski2015understanding} and implemented with Caffe \cite{jia2014caffe} to identify visual features that represent meaningful characteristics of the geometric shape of an object to support grasping.
The network is trained on ImageNet and not retrained for the grasping task.
In this section, we describe how to identify features that activate consistently among the training data. We call these identified features that capture the hierarchical support relations between filters in different CNN layers \textit{hierarchical CNN features}. We discover the locations of these features by repeatedly tracing filter activations backwards in the CNN along a single path. 
We then discuss how grasp points, Cartesian grasp positions of end effectors, are generated based on offsets between features and robot end effectors.

\subsection{Identifying Consistent Features}

Given a set of grasp records that demonstrates grasping objects with similar shapes, our goal is to find a set of visual features that activate consistently. Our assumption is that some of these features represent meaningful geometric information regarding the shape of an object that supports grasping. 
For each observed point cloud, we segment the flat supporting surface using Random Sample Consensus (RANSAC) \cite{fischler1981random} and create a 2D mask that excludes pixels that are not part of the object in the RGB image. This mask is dilated to preserve the boundary of the object. During the forward pass, for each convolution layer we check each filter and zero out the responses that are not marked as part of the object according to the mask. This approach removes filter responses that are not part of the object. The masked forward pass approach is used instead of directly segmenting the object in the RGB image to avoid sharp edges caused by segmentation.

Next, filters that activate consistently over similar object types (\textit{consistent filters}), in the fifth convolution (conv-5) layer are identified. The top $N^5$ filters that have the highest sum of log responses $\sum_{a \in A^5_i} \log a$ among all the grasping examples of the same type of grasp are identified, where $A^5_i$ is the activation map of filter $f^5_i$, and $f^m_i$ represents the $i^{th}$ filter in the $m^{th}$ convolutional layer. In this work, only one type of grasp is demonstrated for the same type of object. We observe that many filters in the conv-5 layer represent high level structures and fire on box-like objects, tube-like objects, faces, etc. Knowing what type of object is observed can determine the type of the grasp but is not sufficient for grasping, since boxes can have different sizes and presented in different pose. However, if lower level features exist such as edges or vertices, robot fingers can be placed relative to them.

\begin{figure}[thb]
\vspace{-1em}
\centering
\includegraphics[width=1.0\linewidth]{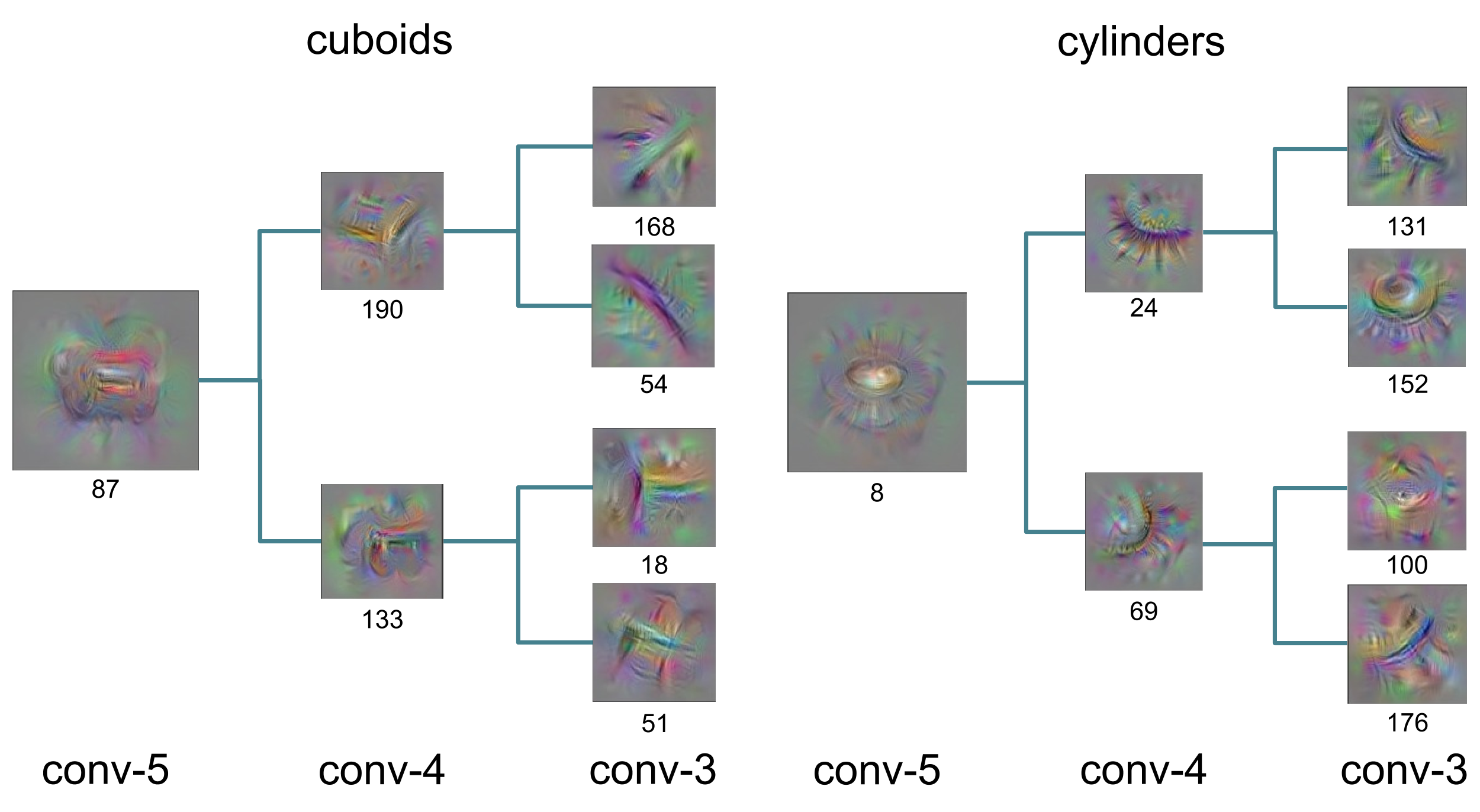}
\caption{Hierarchical CNN feature visualizations among cuboid objects (left) and cylinders (right). Each square figure is the visualization of a CNN filter while the edges connect a lower layer filter to a parent filter in a higher layer. The numbers under the squares are the corresponding filter indices. Filters are visualized using the visualization tool introduced in  \cite{yosinski2015understanding}. Notice that the lower level filters represent local structures of a parent filter.}
\vspace{-0.7em}
\label{fig:cnn_feature}
\end{figure}

CNN features are by nature hierarchical; a filter in a higher layer with little location information is a combination of lower level features with higher spatial accuracy. 
For example, we found that filter $f^5_{87}$ in the conv-5 layer in the CNN represents a box shaped object. This filter is a combination of several filters in the fourth convolution (conv-4) layer. Among these filters, $f^4_{190}$ and $f^4_{133}$ represents the lower right corner and top left corner of the box respectively. Filter $f^4_{190}$ is also a combination of several filters in the third convolution (conv-3) layer. Among these filters, $f^3_{168}$ and $f^3_{54}$ represents the diagonal right edge and  diagonal left edge of the lower right corner of the box.
The activation map of filter $f^3_{168}$ will respond to all diagonal edges in an image, but if we only look at the subset of units of $f^3_{168}$ that has a hierarchical relationship with $f^4_{190}$ in the conv-4 layer and $f^5_{87}$ in the conv-5 layer, local features that correspond to meaningful parts of a box-like object can be identified. Instead of representing a feature with a single filter in the conv-3 layer, our approach uses a tuple of filters to represent a feature, such as ($f^5_{87}$, $f^4_{190}$, $f^3_{168}$) in the previous example.
We call such features \textit{hierarchical CNN features}. Within a hierarchical CNN feature, a higher level filter is called the parent filter of a lower level filter. Figure \ref{fig:cnn_feature} shows the visualization of several hierarchical CNN features identified in cuboid and cylindrical objects, including the features in the previous example.

In order to identify consistent filters in layers lower than the conv-5 layer, the max response of the activation map $A^5_i$ for each of the consistent filters $f^5_i$ in the conv-5 layer is first identified. For each consistent filter $f^5_i$, all responses except for the maximum $a^5_{max}$ are zeroed out and backpropagated to obtain the gradient of the max response with respect to the conv-4 layer: $G^4 = \partial a^5_{max} / \partial \text{conv-4}$. The filters that contribute to $f^5_i$ consistently in the conv-4 layer are identified by picking the top $N^4$ filters that have the highest sum of log gradients $\sum_{g \in G^4_j} \log g$, where $G^4_j$ represents components of the gradient $G^4$ that corresponds to filter $f^4_j$. 
With the same procedure we then find the top $N^3$ filters that contribute to $f^4_j$ consistently by calculating the gradient with respect to the conv-3 layer: $G^3 = \partial g^4_{max} / \partial \text{conv-3}$, where $g^4_{max}$ is the maximum partial derivative of $G^4_j$.
For each hierarchical CNN feature, backpropagation is only performed on one unit per convolutional layer. This process traces backward along a single path recursively and yields a tree structure of hierarchical features.

\begin{figure}[!htbp]
\centering
\includegraphics[width=1\linewidth]{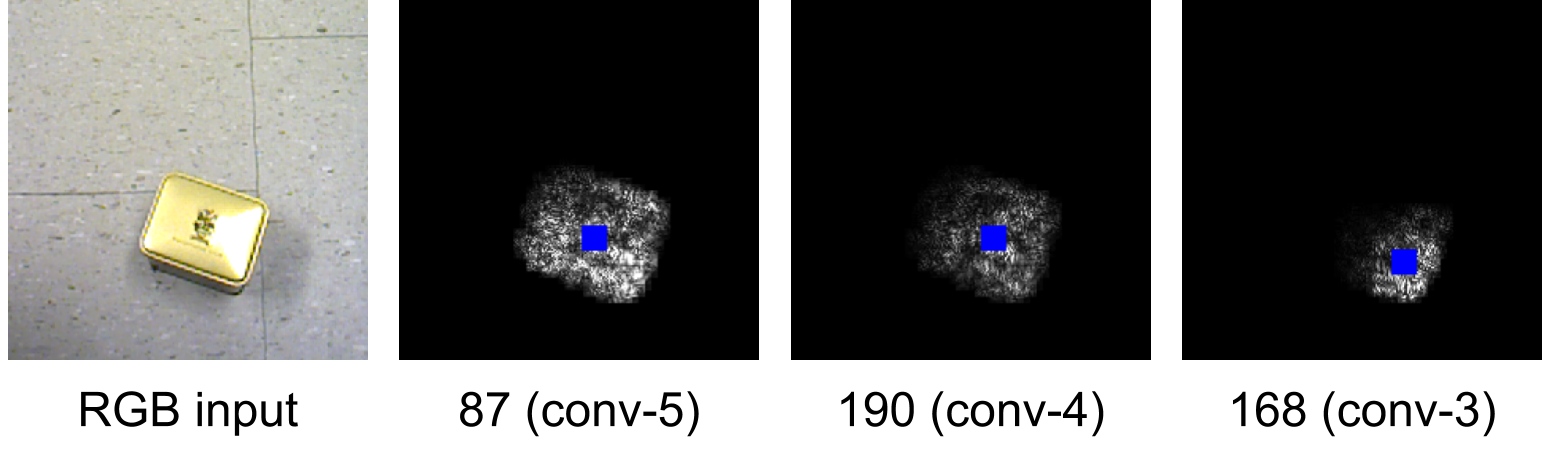}
\caption{Localizing features in different layers. The color image is the input image. The following images from left to right represent the location map of hierarchical CNN features ($f^5_{87}$), ($f^5_{87},f^4_{190}$), and ($f^5_{87},f^4_{190},f^3_{168}$) obtained from backpropagating the feature response along a single path to the image layer. The blue dot in each location map represents the mean of the response locations. The mean response locations of conv-3 layer features are located closer to edges and corners of the cuboid object compared to the conv-4 and conv-5 features. The conv-3 layer features can be interpreted as representing local structures of the cuboid object. }
\label{fig:tbp_result}
\vspace{-0.5em}
\end{figure}

To locate this feature in 3D we then backpropagate to the image and map the mean of the response location to the corresponding 3D point in the point cloud. 
The \textit{response} of a hierarchical CNN feature is defined as the maximum gradient of the lowest layer filter in the tuple, e.g. $g^4_{max}$  for feature ($f^5_{i}$, $f^4_{j}$). Figure \ref{fig:tbp_result} shows an example of results of performing backpropagation along a single path to the image layer from features in the conv-5, conv-4, and conv-3 layers. The conv-3 layer features can be interpreted as representing lower level features such as edges and corners of the cuboid object.

\subsection{Generating Grasp Points}
For each target object we generate grasp points for the index finger, thumb, and the arm. Grasp points are the locations that the corresponding end effectors should be positioned for a successful grasp and are calculated based on a set of relative positions between 3D features and end effectors. 
For each type of object we store a set of possible grasp offsets for each end effector and the corresponding feature. A grasp offset is determined by the relative position from a robot end effector to a feature based on a demonstrated grasp. 
Although modeling a type of grasp based on the offsets between grasp points and feature positions may not result in consistent grasp points when dealing with objects of different sizes at different orientations, there usually exists a subset of feature points that are close to the end effector consistently therefore result in smaller offset variances.
In this work, we associate the robot hand frame to features in the conv-4 layer and the robot index finger and thumb to features in the conv-3 layer. 
This mapping is inspired by the observation that humans are capable of moving their arms toward the object for grasp without recognizing detailed features; the precise locations of low level features are only needed when the finger locations need to be accurate. 


Not all features that fire consistently on the same object type are good features for planning actions. For example, when grasping a box, positioning the index finger, which contacts the back edge of the box, relative to the front edge of the box will result in positions with high variance, since the size of the box may vary. In contrast, positioning the index finger relative to the back edge of the box will result in a lower position variance. For each end effector position we pick the top $N$ hierarchical CNN features that have the lowest 3D position variance among training examples and have the same parent filter in the conv-5 layer. The other features are then removed. The final set of features have the same conv-5 filter $f^5_i$. We found that restricting all the hierarchical CNN features to have the same parent filter allows our approach to perform well in a cluttered scenario. 

During testing, the hierarchical CNN features associated with grasp offsets are first identified. The 3D positions of these features are then located through backpropagation to the 3D point cloud. A set of possible grasp positions are then calculated based on the grasp offsets and the 3D positions of the corresponding hierarchical CNN features. 
The grasp points for the robot hand frame, and end points for the thumb and index finger are then determined by the weighted mean position of the corresponding set of possible grasp positions with the feature responses as weights.
Figure \ref{fig:grasp_eg} shows examples of the grasp points and the set of possible grasp positions on different objects.

\section{EXPERIMENTS}

We ran two sets of experiments. In the first set, we evaluate the difference between the calculated grasp points and the ground truth in the R2 grasping dataset. In the second set, we test our approach by grasping novel objects using Robonaut-2 \cite{diftler2011robonaut} and compare the number of successful grasps with a baseline approach.

\subsection{Experiments on the R2 Grasping Dataset}
In these experiments the performance of our approach on the R2 grasping dataset is analyzed. First, the accuracy of grasp points is measured with cross-validation. Second, approaches with and without the hierarchical CNN features are compared.

\subsubsection{Cross-Validation Results}
\begin{figure}[!htbp]
\centering
\includegraphics[width=1\linewidth]{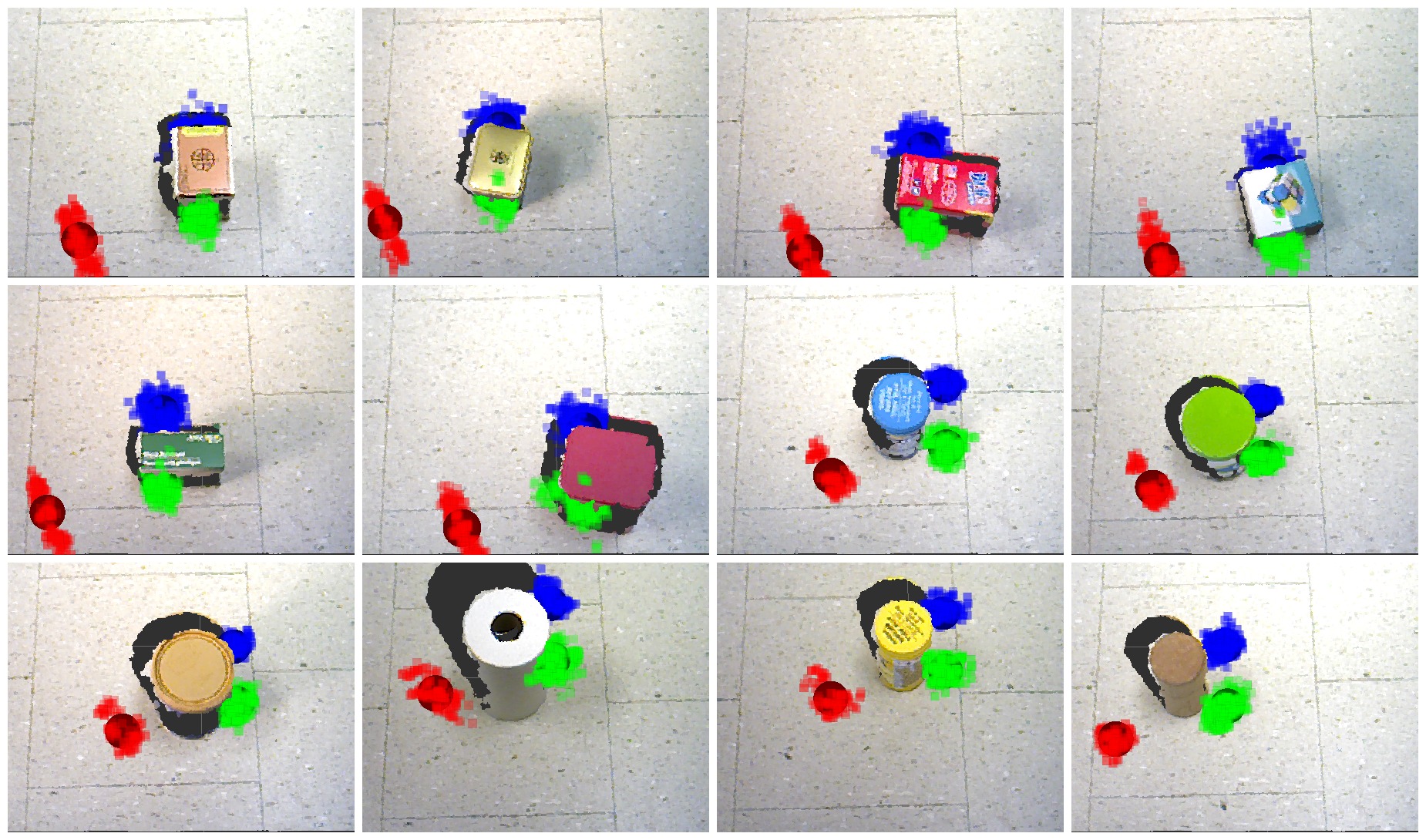}
\caption{Sample cross-validation results for single object scenario. The red, green, and blue spheres represent the calculated grasp points for the hand frame and endpoint positions for the thumb and index finger of the left robot hand. The grasp points are the weighted mean of the colored dots that each represents a possible grasp position based on one training example. Notice that for the cuboid object the grasp points for the thumb and index finger are located on the opposing face and about 3cm away from the left edge of the face as it was trained. For the cylinder object the grasp points for the thumb and index finger are on the right side of the cylinder to form an enveloping grasp. The black pixels are locations behind the point cloud that are not observable.}
\label{fig:grasp_eg}
\end{figure}

Cross-validation is applied by leaving out one object instance at a time during training and testing on the left out object by comparing the calculated grasp points to the ground truth. The distance between the example position and the targeted position of the hand frame, index finger tip, and thumb tip is calculated and shown in Table \ref{table:accuracy_both}. The average grasp position error for the hand frame is higher than the thumb and index finger; this is likely because the positions of local features alone are not sufficient to predict an accurate position for the hand frame. However, since the hand frame is not contacting the object, its position is less crucial for a successful grasp. Figure \ref{fig:grasp_eg} shows a few results of cross-validation on different objects with different pose and lighting. Similar to the training data, the cuboid objects are grasped at positions closer to the left side while the cylindrical objects are grasped such that the fingers would wrap around the cylinder.

\begin{table}[!htbp]   
\centering    
\caption{Average grasp position error on cylindrical and cuboid objects in meters. }  
\footnotesize{                                                                                            
\begin{widetable}{0.47\textwidth}{lllllllr}                                                              
\toprule  
&\multicolumn{7}{c}{cylindrical objects}\\
\cmidrule(l){2-8}
\vspace{-0.3em}  
& cetaphil & wood & maxwell & blue & paper & yellow & \multirow{2}{*}{average} \\
\textit{left hand} & jar & cylinder & can & jar & roll & jar & \\ 
\cmidrule{1-1} \cmidrule(l){2-8}
thumb tip & 0.0197 & 0.0164 & 0.0134 & 0.0202 & 0.0136 & 0.0147 & 0.0163 \\
index tip & 0.011 & 0.0111 & 0.023 & 0.017 & 0.0155 & 0.0126 & 0.015 \\
hand frame      & 0.0128 & 0.0393 & 0.0234 & 0.0183 & 0.0328 & 0.0173 & 0.024\\
\end{widetable} \\
\begin{widetable}{0.47\textwidth}{lllllllr}                                                              
\toprule  
&\multicolumn{7}{c}{cuboid objects}\\
\cmidrule(l){2-8}
\vspace{-0.3em}             
& cube & redtea & bandage & twinings & brillo & tazo & \multirow{2}{*}{average}\\
\textit{left hand} & box & box & box & box & box & box & \\
\cmidrule{1-1} \cmidrule(l){2-8}
thumb tip & 0.0094 & 0.0101 & 0.0093 & 0.0095 & 0.0088 & 0.0111 & 0.01\\
index tip & 0.0135 & 0.0207 & 0.0278 & 0.0134 & 0.0098 & 0.0157 & 0.0168\\ 
hand frame & 0.0241 & 0.0203 & 0.0195 & 0.0177 & 0.0177 & 0.0285 & 0.0213\\
\bottomrule                                                                                                                                                                                                                                                                             
\end{widetable} }                                                                                                                                                               
\label{table:accuracy_both}                                                                            
\end{table}

\subsubsection{Comparison}
To evaluate the proposed hierarchical CNN feature (hier-feat), we compare cross-validation results with four alternative approaches that do not consider relationships between layers. The first approach is a baseline that uses the same set of features identified by the proposed approach but only considers the lowest level filter of the hierarchical CNN features, therefore removing relationship with higher level CNN filters. The grasp points are then generated based on offsets to these individual filters in each layer.
The second approach (indv-filter) also associates grasp points with individual CNN filters instead of hierarchical CNN features but learns the set of filters that fire consistently instead of using the same set of features used by the proposed approach.  Similar to the second approach, the third alternative (conv5-filter) identifies individual CNN filters instead of hierarchical CNN features but only considers filters in the conv-5 layer. The fourth approach (conv5-max) identifies the top five consistent filters in the conv-5 layer and uses the one that has the max response during testing. To make the comparison fair, we also remove filters other than the top $N$ hierarchical CNN features that have the lowest position variance among training examples for the first three comparative approaches. We use $N^5=N^4=N^3=5$ and $N=15$ in this experiment.

The results are shown in the first row of Table \ref{table:tbp_compare_result}. The proposed approach performs better than all four alternatives. However the difference is not significant, this is because the lower level filter that has the maximum response is mostly the same with or without restricting it to have the same parent filter when only one object is presented. In the next test, we show that associating low level filters with high level filters has a greater advantage when low level features may be generated from different high level structures, i.e., when there is clutter present.  The fact that our approach outperforms the conv5-max approach shows the benefit of using lower level features to higher level features on planning actions. In the absence of clutter, the conv5-filter approach performs well because although filters in the conv-5 layer are more likely to represent higher level object structures, many of them also represent low-level features like corners and edges.

\begin{figure}[!htbp]
\centering
\includegraphics[width=1\linewidth]{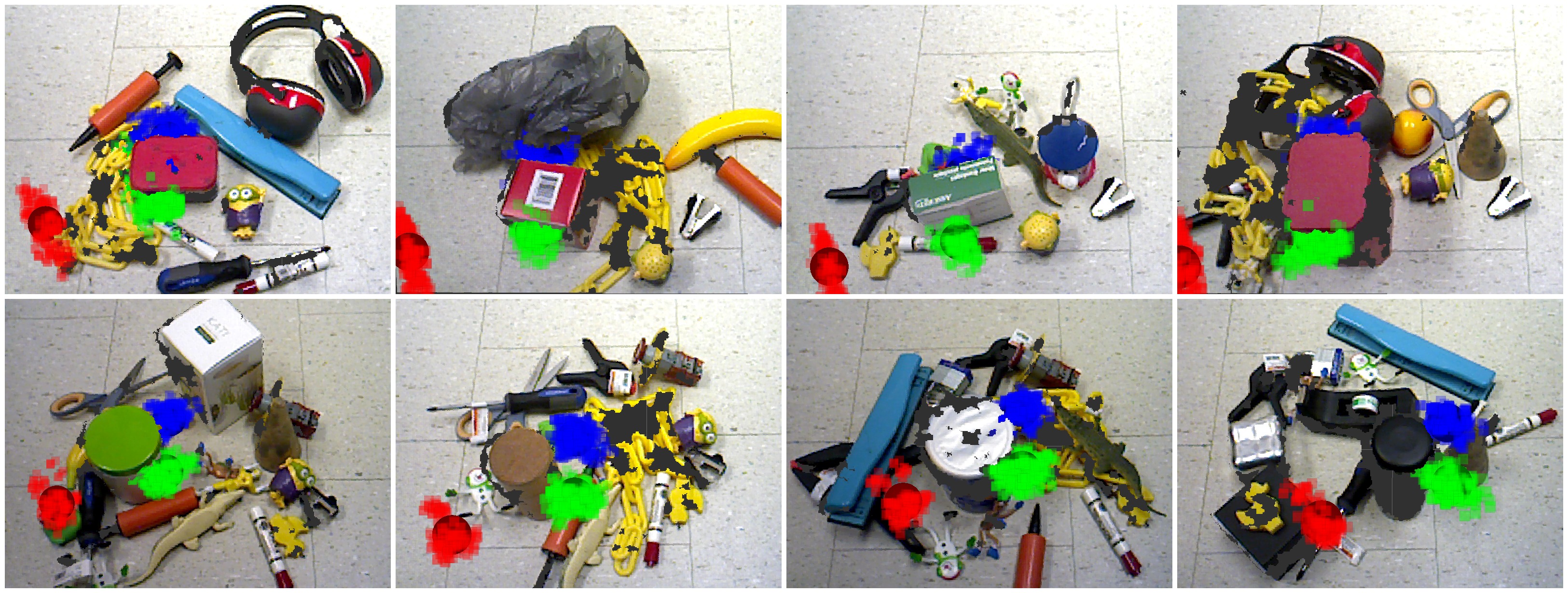}
\caption{Examples of grasping in a cluttered scenario. The red, green, and blue spheres represent the grasp points of the hand frame, thumb tip, and index finger tip of the left robot hand. The grasp points are the weighted mean of the colored dots that each represents a possible grasp position based on one training example. The top row is trained on grasping cuboid objects and the bottom row is trained on grasping cylindrical objects. Notice that our approach is able to identify the only cuboid or cylinder in the scene and generate grasp points similar to the training examples. }
\label{fig:clutter}
\end{figure}

\begin{figure}[!htbp]
\centering
\includegraphics[width=0.85\linewidth]{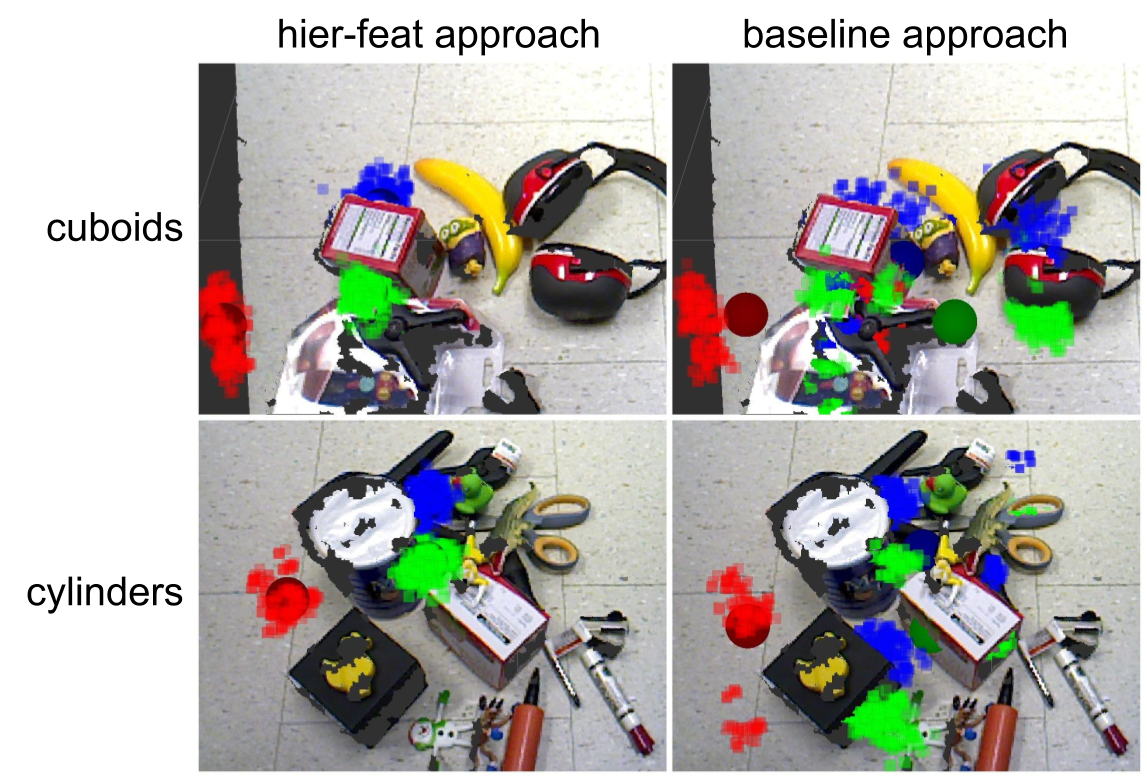}
\caption{Comparison in a cluttered scenario. 
Notice that the colored dots are scattered around in the baseline approach since the highest response filter in conv-3 or conv-4 layer are no longer restricted to the same high level structure.}
\label{fig:tbp_compare}
\vspace{-0.5em}
\end{figure}

Hierarchical CNN features are most useful when the scene is more complex and the same lower layer filter fires at multiple places. Since the proposed approach limits the filters in the conv-3 layer to have the same parent filter in the conv-5 layer, only lower layer features that belong to the same high level structure are considered. These five approaches are further tested on the cluttered test set. A test case is considered successful if the distance errors of the thumb tip and index finger tip are both less than 5cm and the hand frame error is less than 10cm. The results are shown in the second row of Table \ref{table:tbp_compare_result}. Figure \ref{fig:clutter} shows a few example results on cluttered test cases. The proposed approach performs significantly better than the baseline, indv-filter, and conv5-filter approaches since filters may fire on different objects without constraining it to a single high level filter. Figure \ref{fig:tbp_compare} shows two comparison results between the proposed approach and the baseline approach. 

\begin{table}[htbp]   
\vspace{1em}                                                                                            
\centering   
\caption{Comparison on alternative approaches. }     
\footnotesize{                                                                                           
\begin{widetable}{0.47\textwidth}{llllll}                                                              
\toprule     
\vspace{-0.3em}          
& hier- & base- & indv- & conv5- & conv5- \\
& feat &  line & filter & filter & max \\
\midrule
\textbf{Single Object Experiment:} &&&&& \\
cross validation average & 1.719 & 1.805 & 2.114 & 1.755 & 2.138\\
grasp position error (cm) &&&&& \\ 
\midrule
\textbf{Cluttered Experiment:} &&&&& \\
number of failed cluttered  & 2 & 20 & 13 & 19 & 2 \\
cases (24 total) &&&&& \\
\bottomrule
\end{widetable} }                                                                                             
\label{table:tbp_compare_result}                                                                            
\end{table}   

\subsection{Experiments on Robonaut-2}
This section describes the evaluation of the proposed pre-shaping algorithm based on the percentage of successful grasps on a set of novel objects on Robonaut-2 \cite{diftler2011robonaut}. Details on the experimental setting, the hierarchical controller used for pre-shaping, and results are explained.

\subsubsection{Experimental Setting}
For each trial, a single object in the novel object set is placed on a flat surface within the robot's reach. Given the object image and point cloud, the robot moves its wrist and fingers to the pre-shaping pose. After reaching the pre-shaping pose, the hand changes to a pre-defined closed posture and tries to pick up the object by moving the hand up vertically. A grasp is considered to be successful if the object did not drop after the robot tries to pick it up. A total of 100 grasping trials on 10 novel objects with our proposed approach and a comparative baseline approach are tested. 
The novel objects used in this experiment are shown in Figure \ref{fig:dataset}. 

\begin{table}[!htbp]   
\centering    
\caption{Grasp success rate on novel objects based on 5 trials per object. }  
\footnotesize{  
\begin{widetable}{0.47\textwidth}{llllllr}                                                              
\toprule    
&\multicolumn{6}{c}{cylindrical objects}\\
\cmidrule(l){2-7}          
\vspace{-0.3em} 
& \multirow{2}{*}{tumbler} & wipe & basil & hemp & \multirow{2}{*}{duster} & \multirow{2}{*}{average}\\
& & package & container & protein & & \\
\midrule
point cloud & \multirow{2}{*}{40\%} & \multirow{2}{*}{60\%} & \multirow{2}{*}{0\%} & \multirow{2}{*}{20\%} & \multirow{2}{*}{40\%} & \multirow{2}{*}{36\%}\\
approach &&&&&&\\
\midrule
hier-feat & \multirow{2}{*}{100\%} & \multirow{2}{*}{100\%} & \multirow{2}{*}{100\%} & \multirow{2}{*}{100\%} & \multirow{2}{*}{100\%} & \multirow{2}{*}{100\%}\\
(our's) &&&&&&\\
\end{widetable} \\
\begin{widetable}{0.47\textwidth}{llllllr}                                                              
\toprule            
&\multicolumn{6}{c}{cuboid objects}\\
\cmidrule(l){2-7}  
\vspace{-0.3em}            
& cracker & ritz & bevita & bag & energy & \multirow{2}{*}{average}\\
& box & box & box & box & bar box &\\ 
\midrule
point cloud &  \multirow{2}{*}{80\%} &  \multirow{2}{*}{20\%} &  \multirow{2}{*}{60\%} &  \multirow{2}{*}{60\%} &  \multirow{2}{*}{60\%} &  \multirow{2}{*}{52\%}\\
approach &&&&&&\\
\midrule
hier-feat &  \multirow{2}{*}{100\%} &  \multirow{2}{*}{80\%} &  \multirow{2}{*}{100\%} &  \multirow{2}{*}{100\%} &  \multirow{2}{*}{100\%} &  \multirow{2}{*}{96\%}\\
(our's) &&&&&&\\ 
\bottomrule
\end{widetable}                                                                                     
}                                                                                                                                                    
\label{table:grasp_result}                                                                            
\end{table} 

\subsubsection{Hierarchical Controller}

A hierarchical controller constructed from hierarchical CNN features in different CNN layers is implemented to reach the pre-shaping pose.  Given the object image and point cloud, our approach generates targets for the robot hand frame, index finger tip, and thumb tip. The hand frame target is determined based on hierarchical CNN features in the conv-4 layer while the thumb tip and index finger tip target is determined based on hierarchical CNN features in the conv-3 layer. The pre-shaping is executed in two steps. First, the arm controller moves the arm such that the distance from the hand frame to the corresponding grasp point is minimized. Once the arm controller converges, the hand controller moves the wrist and fingers to minimize the sum of squared distances from the index finger tip and thumb tip to their corresponding target. 

These controllers are based on the control basis framework \cite{huber2000hybrid} and can be written in the form $\phi |^{\sigma}_{\tau}$, where $\phi$ is a potential function that describes the sum of squared distances to the targets, $\sigma$ represents sensory resources allocated, and $\tau$ represents the motor resources allocated. In this work, the hand controller is represented as $\mathrm{\phi |^{\sigma_{conv3}}_{\tau_{hand}}}$, where $\mathrm{\tau_{hand}}$ is the hand motor resources and $\mathrm{\sigma_{conv3}}$ is the conv-3 layer hierarchical CNN features; the arm controller is represented as $\mathrm{\phi |^{\sigma_{conv4}}_{\tau_{arm}}}$, where $\mathrm{\tau_{arm}}$ is the arm motor resources and $\mathrm{\sigma_{conv4}}$ is the conv-4 layer hierarchical CNN features.

\subsubsection{Results}
\begin{figure}[!htbp]
\vspace{0.3em} 
\centering
\includegraphics[width=0.95\linewidth]{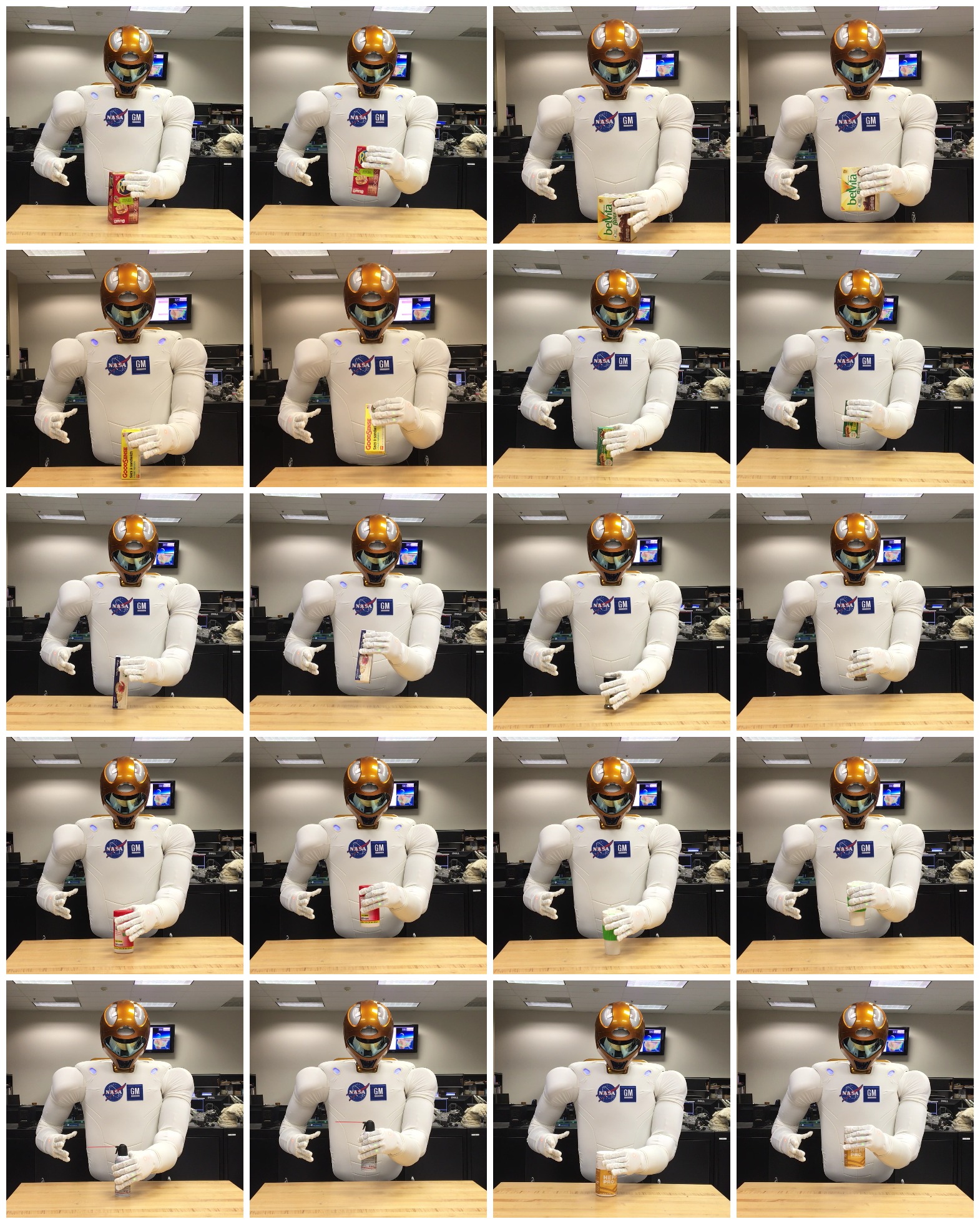}
\caption{Robonaut-2 grasping 10 different novel objects. The first and third columns show the pre-shaping steps while the second and fourth columns show the corresponding grasp and pickup. The cuboid objects are grasped on the faces while the cylinder objects are grasped such that the object is wrapped in the hand.}
\label{fig:grasp_all}
\vspace{-0.8em}
\end{figure}

The proposed algorithm is compared to a baseline point cloud approach that moves the robot hand to a position where the object point cloud center is located at the center of the hand after the hand is fully closed. The results are shown in Table \ref{table:grasp_result}. Among the 50 grasping trials only one grasp failed with the proposed approach due to a failure in controlling the index finger to the target position. This demonstrates that the proposed approach has a much higher probability of success in grasping novel objects than the point cloud based approach. Figure \ref{fig:grasp_all} shows Robonaut-2 grasping novel objects during testing.

\section{DISCUSSION}
Past studies have shown that CNN features are only effective for object recognition when most of the filter responses are used. In \cite{agrawal2014analyzing}, it is shown that to achieve 90\% object recognition accuracy on the PASCAL dataset most object classes require using at least 30 to 40 filters in the conv-5 layer. Research effort to invert CNN features \cite{dosovitskiy2015inverting} has also shown that most information is contained in a population of features with moderate responses and not in the most highly activated filters. In this work, we demonstrate that a subset of hierarchical CNN features that have the same parent filter in the conv-5 layer may be sufficient for controlling the robot end effector for grasping common household objects with simple shapes. Although using all filters may provide subtle information for object classification, we argue that when interacting with objects, the strength of CNN features lies in their hierarchical nature and a small set of filters are sufficient to support manipulation. 

However, the current work is limited to objects with simple shapes that can be represented by a single filter in the conv-5 layer. In future work, we plan to extend to more complicated objects by learning higher layer filters that represent more complicated object types.

\section{CONCLUSION}

In this work, we tackle the problem of pre-shaping a human-like robot hand for grasping based on visual input. 
The hierarchical CNN feature that captures the hierarchical relationship between filters is introduced. The proposed approach first identifies hierarchical CNN features that are active consistently among the same type of grasps and localizes them by backpropagating the response along a single path to the point cloud.
The arm controller is then associated with features in the conv-4 layer while the hand controller is associated with features in the conv-3 layer to perform a grasp. 
The proposed approach is evaluated on the collected dataset and show significant improvement over approaches that do not associate filters in different layers in cluttered scenarios. This solution is further tested in a grasping experiment where a total of 100 grasp trials on novel objects are performed on Robonaut-2. The proposed approach results in a much higher percentage of successful grasps compared to a point cloud based approach.

\section{ACKNOWLEDGMENT}
\footnotesize{
We are thankful to our colleagues Dirk Ruiken, Mitchell Hebert, Michael Lanighan, Tiffany Liu, Takeshi Takahashi, and former members of the Laboratory for Perceptual Robotics for their contribution on the control basis framework code base. We are also grateful to Julia Badger and the Robonaut Team for their support on Robonaut-2.
This material is based upon work supported under Grant NASA-GCT-NNX12AR16A and a NASA Space Technology Research Fellowship. Any opinions, findings, and conclusions or recommendations expressed in this material are those of the authors and do not necessarily reflect the views of the National Aeronautics and Space Administration.
}
\bibliographystyle{IEEEtran}
\bibliography{IEEEabrv,references}

\end{document}